\documentclass[runningheads]{llncs}
\usepackage{graphicx}

\usepackage{tikz}
\usepackage{comment}
\usepackage{amsmath,amssymb} 
\usepackage{color}
\usepackage{array}
\usepackage{arydshln}
\usepackage{multirow}
\usepackage{tabularx}
\usepackage[percent]{overpic} 
\newcolumntype{Y}{>{\centering\arraybackslash}X}

\usepackage[accsupp]{axessibility}  

\usepackage{booktabs}
\usepackage[pagebackref,breaklinks,colorlinks]{hyperref}
\usepackage[capitalize]{cleveref}
\usepackage{xspace}
\usepackage{placeins}
\crefname{section}{Sec.}{Secs.}
\Crefname{section}{Section}{Sections}
\Crefname{table}{Table}{Tables}
\crefname{table}{Tab.}{Tabs.}
\newcommand{\etal}{\textit{et al}.\xspace}

\newcommand{\ie}{\textit{i}.\textit{e}. \xspace}
\newcommand{\eg}{\textit{e}.\textit{g}. \xspace}
\renewcommand{\ie}{\textit{i}.\textit{e}., }
\renewcommand{\eg}{\textit{e}.\textit{g}., }

\newcommand{\BlurPool}{\textit{BlurPool}\xspace}
\newcommand{\MaxPool}{\textit{MaxPool}\xspace}
\newcommand{\UNet}{U-Net\xspace}

\newcommand{\uwCP}{uwCP\xspace}
\newcommand{\uwoCP}{uwoCP\xspace}

\DeclareMathOperator{\model}{f}
\DeclareMathOperator{\replacementmethod}{r}

\DeclareMathOperator{\patch}{\mathbf{X}}

\DeclareMathOperator{\neighborhood}{\mathbf{N}}

\usepackage{caption} 
\usepackage{arydshln}

\usepackage{svg}
\newcommand{\orcid}[1]{\href{https://orcid.org/#1}{\includegraphics[height = 2ex]{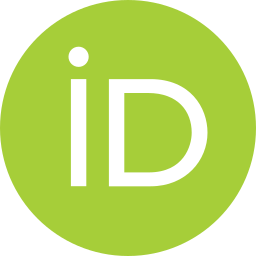}}}
\newcommand{\orcidHT}[1]{\href{https://orcid.org/#1}{\includegraphics[height = 2ex]{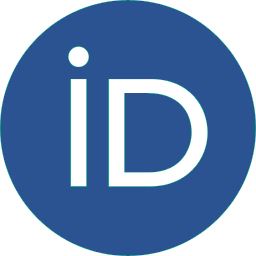}}}
\newcommand{\orcidZEISS}[1]{\href{https://orcid.org/#1}{\includegraphics[height = 2ex]{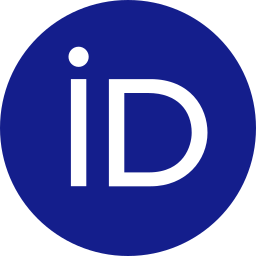}}}
\newcommand{\orcidFMI}[1]{\href{https://orcid.org/#1}{\includegraphics[height = 2ex]{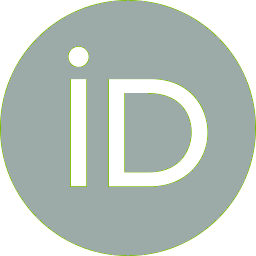}}}
\newcommand\blfootnote[1]{%
  \begingroup
  \renewcommand\thefootnote{}\footnote{#1}%
  \addtocounter{footnote}{-1}%
  \endgroup
}

\begin{document}
\pagestyle{headings}
\mainmatter
\def\ECCVSubNumber{11}  

\title{N2V2 - Fixing Noise2Void Checkerboard Artifacts with Modified Sampling Strategies and a Tweaked Network Architecture}

\titlerunning{N2V2 - Fixing Noise2Void Checkerboard Artifacts}
%
\author{Eva H{\"o}ck\inst{1,\ast}\orcidZEISS{0000-0003-2133-1282} \and
Tim-Oliver Buchholz\inst{2,\ast}\orcidFMI{0000-0001-6953-8915} \and
Anselm Brachmann\inst{1,\ast} \and
Florian~Jug\inst{3,\circledast}\orcidHT{0000-0002-8499-5812} \and
Alexander Freytag\inst{1,\circledast}\orcidZEISS{0000-0002-9041-1334}}
\authorrunning{E. H{\"o}ck \etal}
%
\institute{Carl Zeiss AG, Germany \and
Facility for Advanced Imaging and Microscopy, Friedrich Miescher Institute for Biomedical Research, Basel, Switzerland \and
Jug Group, Fondazione Human Technopole, Milano, Italy
\\
\email{eva.hoeck@zeiss.com, 
tim-oliver.buchholz@fmi.ch, 
anselm.brachmann@zeiss.com,
florian.jug@fht.org,
alexander.freytag@zeiss.com}
}
\maketitle
\blfootnote{$^\ast$, $^\circledast$ Equal contribution}

\begin{abstract}
In recent years, neural network based image denoising approaches have revolutionized the analysis of biomedical microscopy data. 
Self-supervised methods, such as Noise2Void~(N2V), are applicable to virtually all noisy datasets, even without dedicated training data being available.
Arguably, this facilitated the fast and widespread adoption of N2V throughout the life sciences.
Unfortunately, the blind-spot training underlying N2V can lead to rather visible checkerboard artifacts, thereby reducing the quality of final predictions considerably.
In this work, we present two modifications to the vanilla N2V setup that both help to reduce the unwanted artifacts considerably. 
Firstly,
we propose a modified network architecture, \ie using \BlurPool instead of \MaxPool layers throughout the used \UNet, rolling back the residual-\UNet to a non-residual \UNet, and eliminating the skip connections at the uppermost \UNet level.
Additionally,
we propose new replacement strategies to determine the pixel intensity values that fill in the elected blind-spot pixels.
We validate our modifications on a range of microscopy and natural image data. 
Based on added synthetic noise from  multiple noise types and at varying amplitudes, we show that both proposed modifications push the current state-of-the-art for fully self-supervised image denoising.
\end{abstract}

\section{Introduction}
\label{sec:introduction}
Fluorescence microscopy is one of the major drivers for discovery in the life sciences.
The quality of possible observations is limited by the optics of the used microscope, the chemistry of used fluorophores, and the maximum light exposure tolerated by the imaged sample. 
This necessitates trade-offs, frequently leading to rather noisy acquisitions as a consequence of preventing ubiquitous effects such as photo toxicity and/or bleaching.
While the light efficiency in fluorescence microscopy can be optimized by specialized hardware, \eg by using Light Sheet or Lattice Light Sheet microscopes, software solutions that restore noisy or distorted images are a popular additional way to free up some of the limiting photon budget. 

\begin{figure}[t]
     \centering
    \includegraphics[width=0.9\textwidth]{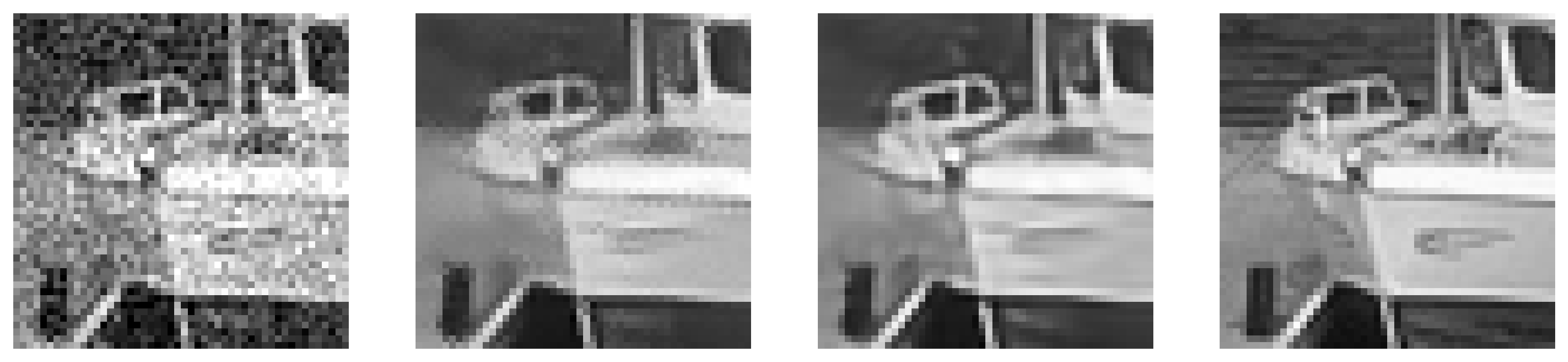}
     \begin{tabularx}{0.95\textwidth}{Y Y Y Y}
     Input & N2V~\cite{krull2019noise2void} & \textbf{N2V2 (Ours)} & GT
     \end{tabularx}        
    \caption{Self-supervised denoising of noisy data~(left). 
    Results obtained with Noise2Void~(N2V)~\cite{krull2019noise2void} (here shown without residual connection and with sampling without the center point) are subject to clearly visible checkerboard artifacts (2\textsuperscript{nd} column).
    Our proposed method, Noise2Void~v2~(N2V2), visibly reduces these artifacts, leading to improved quality results (3\textsuperscript{rd} column, here shown with median center pixel replacement).
    The last column shows ground truth (not available to either method)}
    \label{fig:teaser}
\end{figure}
Algorithmic image restoration is the reconstruction of clean images from corrupted versions as they were acquired by various optical systems. 
A plethora of recent work shows that CNNs can be used to build powerful content-aware image restoration~(CARE) methods~\cite{Weigert2017,weigert2018content,zhang2017beyond,zhang2019making,lehtinen2018noise2noise,buchholz2019content}. 
However, when using supervised CARE approaches, as initially proposed in~\cite{weigert2018content}, pairs of clean and distorted images are required for training the method.
For many applications in the life-sciences, imaging such clean ground truth data is either impossible or comes at great extra cost, often rendering supervised approaches as being practically infeasible~\cite{krull2019noise2void}.

Hence, self-supervised training methods like Noise2Void~(N2V) by Krull \etal~\cite{krull2019noise2void}, which operate exclusively on single noisy images, are frequently used in  life-science research~\cite{batson2019noise2self,krull2019noise2void,laine2019high,pn2v,lalit2020ppn2v}. 
Such \emph{blind-spot approaches} are enabled by excluding/masking the center (blind-spot) of a network's receptive field and then training the network to predict the masked intensity. 
These approaches collectively assume that the noise to be removed is pixel-wise independent (given the signal) and that the true intensity of a pixel can be predicted after learning a content-aware prior of local image structures from a body of noisy data~\cite{krull2019noise2void}.

More recently, methods that can sample the space of diverse interpretations of noisy data were introduced~\cite{divnoising,hdn}. 
While these approaches show great performance on denoising and even artifact removal tasks, the underlying network architectures and training procedures are space and time demanding~\cite{hdn} and can typically not be used on today's typical consumer workstations and laptops. 
Hence, comparatively small blind-spot networks like N2V are available via consumer solutions such as \textit{Fiji}~\cite{Schroeder2021-qt,Buchholz2020-vd}, \textit{ZeroCostDL4Mic}~\cite{zero}, or the \textit{BioImage.IO Model Zoo}~\cite{Ouyang2022-ModelZoo}, and are therefore still the most commonly used self-supervised denoising methods.

Still, one decisive problem with blind-spot approaches such as N2V is that  checkerboard artifacts can commonly be observed (see \Cref{fig:teaser} for an illustrative example). 
Hence, in this work we present Noise2Void~v2~(called N2V2), a variation of N2V that addresses the problem with checkerboard artifacts by a series of small but decisive tweaks.

More concretely, the contributions of our work are: 
$(i)$~showcasing and inspecting the short-comings of N2V,
$(ii)$~proposal of an adapted \UNet architecture that replaces \textit{max-pooling} layers with \textit{max-blur-pooling}~\cite{zhang2019making} layers and omits the top-most skip-connection,
$(iii)$~introduction of blind-spot pixel replacement strategies,
$(iv)$~a systematic evaluation of our proposed replacement strategies and architectural changes on the BSD68 dataset~\cite{krull2019noise2void}, the Mouse, Convallaria and Flywing datasets from~\cite{divnoising,lalit2020ppn2v} and two newly added salt and pepper (S\&P) noise regimes, and
$(v)$~proposal of a new variation on the Convallaria dataset from~\cite{lalit2020ppn2v} that addresses what we believe to be non-ideal setup choices.

\section{Related Work}
\label{sec:related_work}
The original CARE work by Weigert~\etal~\cite{weigert2018content} steered our field away from more established and non-trained denoising methods towards modern data-driven deep denoising methods.
With these fully supervised methods deep neural networks are trained on pairs of low-quality and high-quality images that are pixel-perfectly aligned and contain the exact same objects (or `scene'). 

Such pairs need to be carefully acquired at the microscope, typically by varying acquisition parameters such as exposure time and illumination intensity.
In certain modalities, \eg cryo transmission electron microscopy (cryo-TEM), acquisition of high-exposure images is impossible and even the acquisition of pairs of noisy images is undesirable~\cite{buchholz2019content}.

However, if pairs of independently noisy images are available, Noise2Noise (N2N) training~\cite{lehtinen2018noise2noise} can be applied and high quality predictions are still achievable.
Later, Buchholz~\etal~\cite{buchholz2019cryo}, extended these ideas to full cryo electron tomography (cryo-ET) workflows~\cite{Jimenez_de_la_Morena2022-ax}.

Still, clean ground truth data or a second set of independently noisy images is typically not readily available.
This motivated the introduction of self-supervised methods such as Noise2Void~\cite{krull2019noise2void} and Noise2Self~\cite{batson2019noise2self}.
The simplicity and applicability of these methods makes them, to-date, the de-facto standard approach used by many microscopists on a plethora of imaging modalities and biological samples.
All such blind-spot approaches exploit the fact that for noise which is independent per pixel (given the signal), the intensity value of any given pixel can in principle be estimated from examining the pixels image context (surrounding). 
This is precisely what content-aware image restoration approaches do.
Pixel-independent noise, instead, can by definition not be predicted, leading to a situation where the loss minimizing prediction does, in expectation, coincide with the unknown signal at the predicted pixel~\cite{krull2019noise2void,batson2019noise2self,lehtinen2018noise2noise}.

An interesting extension of N2V was introduced by Krull~\etal~\cite{pn2v}.
Their method, called \textit{Probabilistic Noise2Void}~(PN2V), does not only predict a single (maximum likelihood) intensity value per pixel, but instead an entire distribution of plausible pixel intensity values (prior).
Paired with an empirical (measured) noise-model~\cite{pn2v,lalit2020ppn2v}, \ie the distributions of noisy observations for any given true signal intensity (likelihood), PN2V computes a posterior distribution of possible predicted pixel intensities and returns, for example, the minimum mean squared error~(MMSE) of that posterior.

A slightly different approach to unsupervised image denoising was proposed by Prakash~\etal~\cite{divnoising,hdn}.
Their method is called (Hierarchical) DivNoising and employs a variational auto-encoder (VAE), suitably paired with a noise model of the form described above~\cite{pn2v,lalit2020ppn2v}, that can be used to sample diverse interpretations of the noisy input data.

In contrast to these probabilistic approaches, we focus on N2V in this work due to its popularity. Hence, we aim at making a popular method, which is at the same time powerful \emph{and} simple, even more powerful.

\subsection*{Particularities of the Publicly Available Convallaria Dataset}
\label{subsec:conva_data}
Self-supervised denoising methods are built to operate on data for which no high-quality ground truth exists.
This makes them notoriously difficult to evaluate quantitatively, unless when applied on data for which ground truth is obtainable.

To enable a fair comparison between existing and newly proposed methods, several benchmark datasets have been made available over the years. 
One example is the \textit{Convallaria} data, first introduced by Lalit~\etal~\cite{lalit2020ppn2v}.
This dataset consists of $100$ noisy short exposure fluorescence acquisitions of the same $1024 \times 1024$px field of view of the same sample. 
The corresponding ground truth image used to compare against was created by pixel-wise averaging of these $100$ independently noisy observations.

In later work~\cite{lalit2020ppn2v,divnoising}, the proposed methods were trained on $95$ of the individual noisy images, while the remaining $5$ images were used for validation purposes.
For the peak signal-to-noise ratio (PSNR) values  finally reported in these papers, the predictions of the top left $512 \times 512$ pixels of all 100 noisy images were compared to the corresponding part of the averaged ground truth image. 
In this paper we refer to this dataset and associated train/validation/test sets as  \textit{Convallaria\_95}.

\begin{figure}[b]
    \centering
    \includegraphics[width=0.95\textwidth, trim={0 0 10 0},clip]{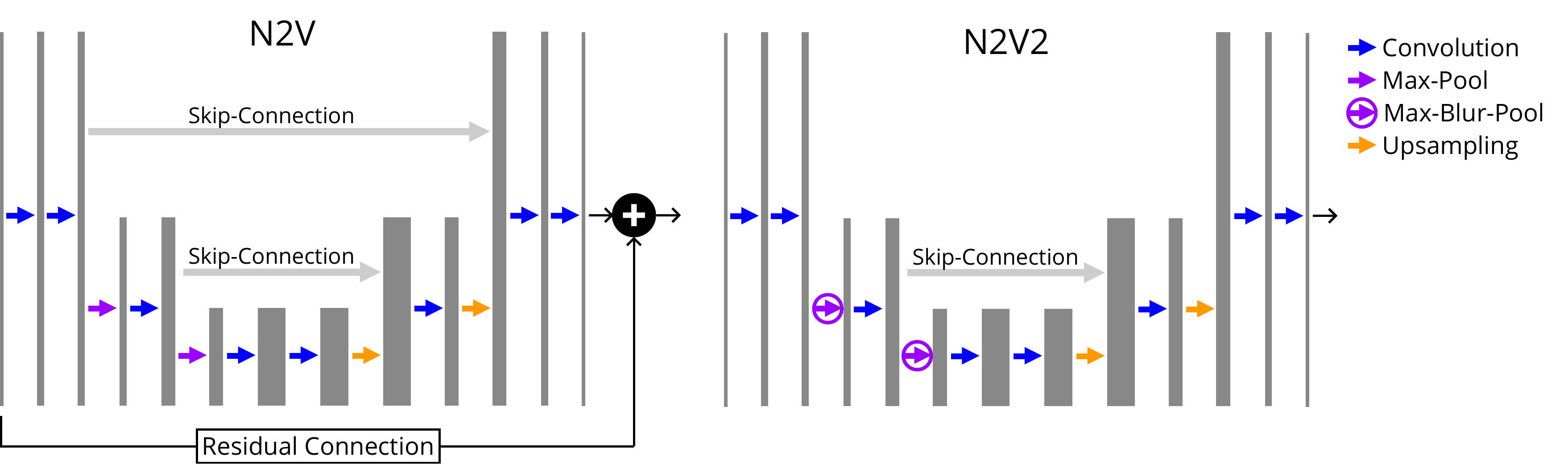}
    \caption{\emph{Left:} the N2V architecture in \cite{krull2019noise2void} is a standard \UNet~\cite{ronneberger2015u} with a residual connection.
    \emph{Right:} 
    our N2V2 architecture drops the residual connection, removes the top-most skip-connection and replaces the max-pooling layers with max-blur-pooling layers~\cite{zhang2019making}}
    \label{fig:n2v2_architecture}
\end{figure}

We are convinced that training self-supervised image denoising methods on $95$ noisy observations of the exact same field of view is leading to slightly misleading results (that overestimate the performance to be expected from the tested method) in cases where only one noisy image per sample exists.
Also note that in cases where already as few as two noisy observations per sample are available, a network can be trained via N2N~\cite{lehtinen2018noise2noise}.
With $95$ such instances available, one could even average those and use the average as ground truth for fully supervised CARE training~\cite{weigert2018content}.

Hence, we propose here to use the Convallaria data differently, namely by selecting one of the $100$ images and tiling it into $64$ tiles of $128\times 128$px. 
Of these tiles, $56$, $4$, and $4$ are then used for training, validation, and testing respectively. See the supplementary material for more information. 
We refer to this data and train/validation/test split as \textit{Convallaria\_1}
Please see \Cref{sec:evaluation} for a thorough evaluation of achievable denoising results when using \textit{Convallaria\_95} versus \textit{Convallaria\_1}.

\section{Method}
\label{sec:method}
As can be seen in \cref{fig:teaser}, denoising predictions from a vanilla N2V model can exhibit considerable amounts of unwanted checkerboard artifacts.
After observing this phenomenon on several datasets, our hypothesis is that these artifacts are caused by two aspects in the vanilla N2V design: 
$(i)$~missing high-frequency suppression techniques to counteract strongly noisy pixel values that really stick out with respect to their close neighbors, and 
$(ii)$~an amplification of this effect due to N2V's self-supervised input replacement scheme (blind-spots). 
Below, we describe two measures which we introduce in N2V2 to mitigate these problems.

\subsection{A Modified Network Architecture for N2V2}
\label{subsec:architecture_changes}
The default N2V configuration employs a \textit{residual} \UNet with $2 \times 2$ max-pooling layers throughout the encoder~\cite{krull2019noise2void}. 
We propose to change this architecture in three decisive ways by
$(i)$~removing the residual connection and instead use a regular \UNet,
$(ii)$~removing the top-most skip-connection of the \UNet to further constrain the amount of high-frequency information available for the final decoder layers, and
$(iii)$~replacing the standard max-pooling layers by max-blur-pool layers~\cite{zhang2019making} to avoid aliasing-related artifacts.
In \Cref{fig:n2v2_architecture}, we highlight all proposed architectural changes which we propose for N2V2.

\begin{figure}[bt]
    \centering
    \begin{overpic}[width=0.7\linewidth,trim=0 0 140 0,clip]{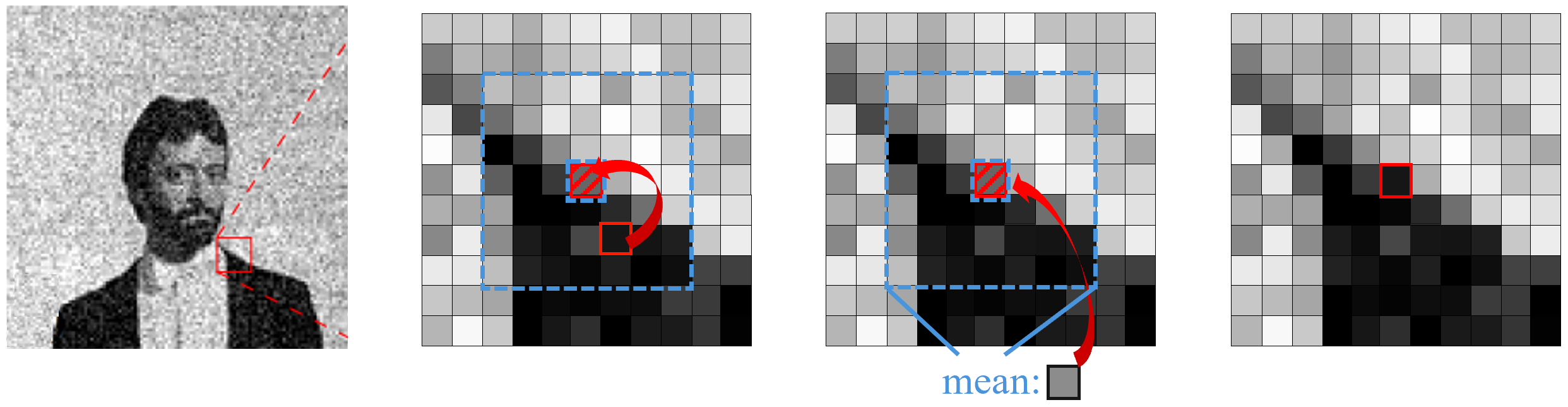}
    \put(36,0.7){\small{(a)}}
    \put(70,0.7){\small{(b)}}
    \end{overpic}
    \caption{Pixel replacement strategies are key for efficient N2V training. \textbf{(a)} the original N2V replacement strategy in~\cite{krull2019noise2void} chooses a random pixel from the center pixel's local neighborhood, which may lead to artifacts like checkerboard patterns in denoised images (see \Cref{fig:teaser}). \textbf{(b)} Our proposed average center pixel replacement strategy calculates the mean or median of the pixel's local neighborhood while excluding the center pixel itself}
    \label{fig:replacement}
\end{figure}

\subsection{New Sampling Strategies to Cover Blind-Spots}
\label{subsec:replacement_methods}
As mentioned before, self-supervised denoising methods introduce blind-spots, effectively asking the network to perform content-aware single pixel inpainting~\cite{krull2019noise2void,batson2019noise2self,pn2v,lalit2020ppn2v}.

During training, a self-supervised loss is employed that compares measured (and hence left out) pixel values with the corresponding pixel values predicted by the trained network (to which only the local neighborhood of the respective blind-spot pixels is given). 

Let $\patch \in \mathfrak{N}^{ w \times h}$ be a patch in a given input image with intensity range $\mathfrak{N}$. Without loss of generality, let $x_{i}$ be a single pixel in $\patch$.
As loss for a given patch $\patch$, N2V~\cite{krull2019noise2void} starts with proposing
\begin{equation}
    \label{eq:n2v_loss_naive}
    \mathfrak{L}_{\text{N2V-na\"ive}}
         \bigl( \patch, \model\left(\cdot\right) \bigr)
    =
    \bigl(  \model\left( \patch \backslash \{x_{i}\} \right)_{i} - x_{i} \bigr)^2 \quad,
\end{equation}
where $\patch \backslash \{x_{i}\}$ denotes the exclusion of pixel $x_i$ from $\patch$.
This exclusion operation would be computational inefficient when implemented na\"ively in convolutional networks.
Krull~\etal have therefore proposed not to \emph{exclude} $x_{i}$, but rather to \emph{replace} $x_{i}$'s value and thereby hiding the true intensity of blind-spot pixels:
\begin{equation}
    \label{eq:n2v_loss}
    \mathfrak{L}_{\text{N2V}}
         \bigl( \patch, \model\left(\cdot\right) \bigr)
    =
    \bigl(  \model\left( \replacementmethod\left(\patch\right) \right)_{i} - x_{i} \bigr)^2 \quad,
\end{equation}
where $\replacementmethod\left(\patch\right)$ assigns a new value to $x_{i}$ in $\patch$.
While \cref{eq:n2v_loss} can be evaluated efficiently compared to \cref{eq:n2v_loss_naive}, 
it turns out that the choice of  $\replacementmethod\left(\patch\right)$ is more sensitive than originally believed, with some choices leading to emphasized visual artifacts like the ones shown in \Cref{fig:teaser}.

\subsubsection{Default N2V Pixel Sampling Strategies (\uwCP and \uwoCP).}
In \cite{krull2019noise2void}, Krull~\etal analyze different blind-spot pixel replacement methods and settle for one default method in their public implementation\footnote{\url{https://github.com/juglab/n2v}}.
This default method, called UPS in N2V, is ubiquitously used by virtually all users world-wide and samples a pixel $x_j$ uniformly at random from a small neighborhood $\neighborhood \subset \patch$ of size $w' \times h'$ around a blind-spot pixel $x_i$ (including $x_i$ itself).
We refer to this replacement technique as \uwCP, and illustrate it in \Cref{fig:replacement}. 

The first obvious observation is that with probability of $1 / \left(w'\cdot h'\right)$, $j$ will be equal to $i$, \ie no replacement is happening.
In these cases, the best solution to any model $\model\left(\cdot\right)$ will be the identity, which is clearly not intended for denoising tasks.
Therefore, in PN2V~\cite{pn2v}, the available implementation\footnote{\url{https://github.com/juglab/pn2v}} started using a slightly altered sampling strategy that excludes the center pixel from being sampled, \ie $i \neq j$, which we refer to as \uwoCP.

\subsubsection{Blind-Spot Replacement Strategies for N2V2.}
In contrast to the blind-spot replacement strategies via sampling from $\neighborhood$, we propose to compute replacement strategies computed from the entire pixel neighborhood $\neighborhood$ (but without $x_{i}$) .
Specifically, we propose
$\replacementmethod_{\text{mean}}(\neighborhood) = \operatorname{mean}(\neighborhood  \backslash \{x_{i}\} )$
and 
$\replacementmethod_{\text{median}}(\neighborhood) = \operatorname{median}(\neighborhood  \backslash \{x_{i}\} )$
as replacement strategies, and refer to them as \textit{mean} and \textit{median} replacement strategies, respectively.

Note that the exclusion of the center pixel is important
in order to fully remove any residual information about the blind-spot pixels to be masked.
Please refer to \Cref{fig:replacement} for a visual illustration.

\section{Evaluation}
\label{sec:evaluation}
We evaluate our proposed pixel replacement strategies and the architectural changes on multiple datasets and perform different ablation studies.
The covered datasets with their experiment details are described in \Cref{subsec:datasets}.
Evaluation metrics are listed in \Cref{subsec:evaluation_metrics}.
Results on data with S\&P noise are given in \Cref{subsec:sp_results}.
Complementary results with other noise types are given in \Cref{subsec:results_gaussian_noise}.
In \Cref{subsec:results_convallaria}, we finally shed light on aspects of generalization and evaluation in scenarios where only single noisy recordings are available.

\subsection{Dataset Descriptions and Training Details}
\label{subsec:datasets}
All dataset simulation and method evaluation code, together with the used training configurations, is publicly available on GitHub\footnote{\url{https://github.com/fmi-faim/N2V2_Experiments}}.
The N2V2 code is part of the official Noise2Void implementation\footnote{\url{https://github.com/juglab/n2v}}.

\subsubsection{General Settings}
All our hyper-parameter choices are based on previous publications~\cite{krull2019noise2void,pn2v,hdn} to keep the reported values comparable.
Hence, we used an Adam optimizer with a reduce learning rate on plateau scheduler (patience of $10$), chose $0.198\%$ random pixels per training patch as blind-spots and the pixel replacement was performed within a neighborhood of size $w' = h' = 5$. 

\subsubsection{BSD68}
An evaluation on natural images is done with the BSD68 dataset as used in the original N2V paper~\cite{krull2019noise2void}.
For training,
we use the same $400$ natural gray scale images of size $180 \times 180$px from~\cite{zhang2017beyond}.
From those, 
$396$ are used as training data and $4$ for validation as described in N2V.
BSD68 networks are of depth $2$ with $96$ initial feature maps and are trained for $200$ epochs, with $400$ steps per epoch, a batch size of $128$,
and an initial learning rate of $0.0004$.

\begin{table}[tb]
    \centering
     \begin{tabular}{l l@{\hskip 12pt} c@{\hskip 12pt} c@{\hskip 12pt} c} 
        \toprule
          & \textbf{Method} & \textbf{Mouse SP3} & \textbf{Mouse SP6} & \textbf{Mouse SP12} \\
          \midrule
          & Input                                & $20.03$ & $18.72$ & $17.76$ \\
         \hdashline
         \multirow{8}{*}{\shortstack{Fully \\self- \\ supervised}}
          & N2V as in \cite{krull2019noise2void} & $21.32$ & $20.69$ & $20.99$ \\
          & N2V w/ \uwoCP as in \cite{pn2v}      & $35.17$ & $34.24$ & $33.49$ \\
          & N2V w/o res, w/ \uwoCP               & $35.44$ & $34.89$ & $34.12$ \\
           \cdashline{2-5}
          & N2V w/o res  w/ mean                & $35.29$ & $34.71$ & $33.66$ \\
          & N2V w/o res  w/ median              & $35.23$ & $35.07$ & $33.45$ \\
          & N2V2 w/ \uwCP                        & $35.74$ & $35.32$ & $34.19$ \\
          & N2V2 w/ \uwoCP                       & $\mathbf{35.91}$ & $35.47$ & $34.52$ \\
          & N2V2 w/ mean                         & $35.51$ & $35.01$ & $34.17$ \\
          & N2V2 w/ median                       & $35.81$ & $\mathbf{35.50}$ & $\mathbf{34.54}$ \\
         \hdashline
         \multirow{2}{*}{\shortstack{Self-\\ supervised}} 
         & PN2V \cite{pn2v}                      & $29.67$ & N/A & N/A \\
         & DivNoising \cite{divnoising}          & $36.21$ & N/A & N/A \\
         \hdashline
         Supervised 
         & CARE \cite{weigert2018content}  & $\underline{37.03}$ & N/A & N/A \\
         \bottomrule
     \end{tabular}
     \caption{Quantitative results on data with simulated S\&P noise. Results are given in dB of averaged PSNR on test data. Overall best is $\text{\underline{underlined}}$. Best fully self-supervised is in \textbf{bold}}
     \label{tab:s&presults}
\end{table}
\begin{table}[tb]
    \centering
     \begin{tabular}{l l@{\hskip 12pt} c@{\hskip 12pt} c@{\hskip 12pt} c} 
         \toprule
                  & \textbf{Method} & \textbf{Flywing G70} & \textbf{Mouse G20} & \textbf{BSD68} \\
         \midrule
          & Input                     & $17.67$ & $22.52$ & $21.32$ \\
         \hdashline
         \multirow{10}{*}{\shortstack{Fully \\self- \\ supervised}} &
           N2V as in \cite{krull2019noise2void}   & $25.20$ & $34.12$ & $27.70$ \\
          & N2V w/ \uwoCP as in \cite{pn2v}       & $25.04$ & $33.94$ & $27.37$ \\
          & N2V w/o res, w/ \uwoCP                & $25.24$ & $34.20$ & $26.95$ \\
          \cdashline{2-5}
          & N2V w/o res  w/ mean         & $25.54$ & $34.49$ & $28.25$ \\
          & N2V w/o res  w/ median       & $\mathbf{25.57}$ & $34.41$ & $27.49$ \\
          & N2V w/ bp w/ \uwCP            & $25.30$ & $34.17$ & $27.69$ \\
          & N2V w/o sk w/ \uwCP           & $25.49$ & $34.63$ & $27.88$ \\
          & N2V2 w/ \uwCP                 & $25.42$ & $34.65$ & $28.04$ \\
          & N2V2 w/ \uwoCP                & $25.49$ & $34.59$ & $27.97$ \\
          & N2V2 w/ mean                  & $25.48$ & $34.61$ & $28.31$ \\
          & N2V2 w/ median                & $25.46$ & $\mathbf{34.74}$ & $\mathbf{28.32}$ \\
         \hdashline
         \multirow{2}{*}{\shortstack{Self-\\ supervised}} & PN2V \cite{pn2v} & $24.85$ & $34.19$ & N/A \\
          & DivNoising \cite{divnoising} & $25.02$ & $34.13$ & N/A \\
         \hdashline
         Supervised & CARE \cite{weigert2018content} & $\underline{25.79}$ & $\underline{35.11}$ & $\underline{29.06}$ \\
         \bottomrule
     \end{tabular}
     \caption{Quantitative results: simulated Gaussian noise. Results are given in dB of averaged PSNR on test data. Overall best is $\text{\underline{underlined}}$. Best fully self-supervised is in \textbf{bold}}
     \label{tab:gaussiannoiseresults}
\end{table}

\subsubsection{Convallaria}
We evaluate on the fluorescence imaging dataset Convallaria by~\cite{lalit2020ppn2v}.
Due to its specialities as described in \Cref{subsec:conva_data},
we call it Convallaria\_95.
Additionally, we introduce the Convallaria\_1 dataset where the input corresponds to only one single noisy observation of $1024 \times 1024$px and the corresponding ground truth is the average of the $100$ noisy Convallaria observations.
This image pair is divided into non-overlapping patches of $128 \times 128$px, resulting in $64$ patches.
These patches are shuffled and $56$, $4$, and $4$ patches are selected as training, validation and test data respectively (see Supplementary Figure~S3).
We train Convallaria\_95 and Convallaria\_1 networks with depth $3$,  with $64$ initial feature maps,  and  for $200$ epochs, with $10$ steps per epoch, a batch size of $80$, and an initial learning rate of $0.001$.

\subsubsection{Mouse}
We further conduct evaluations based on the ground truth Mouse dataset from the DenoiSeg paper~\cite{buchholz2020denoiseg},
showing cell nuclei in
the developing mouse skull.
The dataset consists of $908$ training and $160$ validation images of size $128 \times 128$px, with another $67$ test images of size $256 \times 256$px. 
From this data,
we simulate Mouse\_G20 by adding Gaussian noise with zero-mean and standard deviation of $20$.
Furthermore,
we simulate Mouse\_sp3, Mouse\_sp6 and Mouse\_sp12, three datasets dominated by S\&P noise.
More specifically,
we apply Poisson noise directly to the ground truth intensities, 
then add Gaussian noise with zero-mean and standard deviation of $10$, 
and clip these noisy observations to the range $[0, 255]$.
Then,
we randomly select $p$\%  of all pixels ($p \in [3, 6, 12]$) and set them to either $0$ or $255$ with a probability of $0.5$.
We train networks on the Mouse dataset with depth $3$,
with $64$ initial feature maps,
and for $200$ epochs, with $90$ steps per epoch, a batch size of $80$ and an initial learning rate of $0.001$.

\subsubsection{Flywing}
Finally, we report results on the Flywing dataset from the DenoiSeg~\cite{buchholz2020denoiseg}, showing membrane labeled cells in a flywing.
We follow the data generation protocol described in~\cite{divnoising}, \ie we add zero-mean Gaussian noise with a standard deviation of $70$ to the clean recordings of the dataset.
The data consists of $1\,428$ training and $252$ validation patches of size $128 \times 128$px, with additional $42$ images of size $512 \times 512$px for testing.
On the flywing dataset, we train networks with of depth $3$, 
with $64$ initial feature maps,
and for $200$ epochs, with $142$ steps per epoch, a batch size of $80$ and an initial learning rate of $0.001$.

\subsubsection{Data Augmentation}
All training data is $8$-fold augmented by applying three $90\deg$ rotations and flipping. 
During training,
random $64 \times 64$ crops are selected from the provided training patches as described in~\cite{krull2019noise2void}.

\subsection{Evaluation Metrics}
\label{subsec:evaluation_metrics}

\begin{figure}[tb]
     \centering
     \begin{tabularx}{0.9\textwidth}{Y Y Y Y Y Y Y}
     Input & \shortstack{N2V \\ w/o res\\ w/ \uwoCP}  & \shortstack{N2V\\ w/o res,\\ w/ mean} & \shortstack{N2V\\ w/o res,\\ w/ median} & \shortstack{N2V2\\ w/ median}  & GT \\ 
     \hline
     \end{tabularx}
    \includegraphics[width=0.9\textwidth]{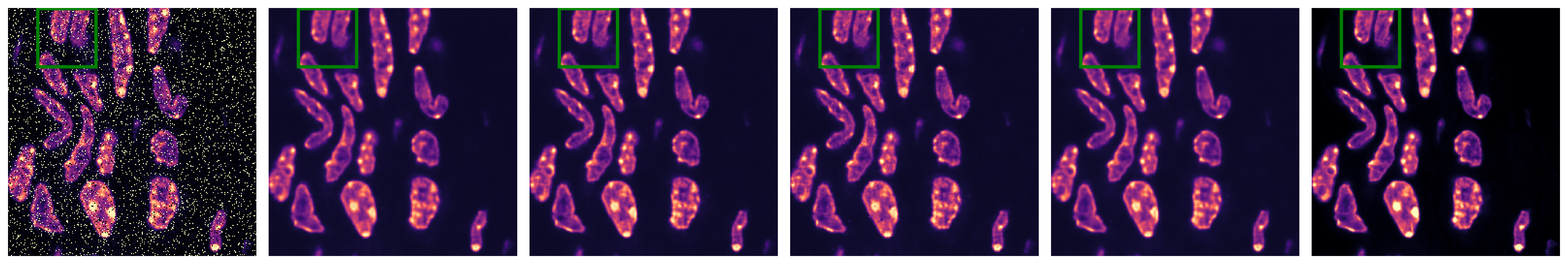}
    \includegraphics[width=0.9\textwidth]{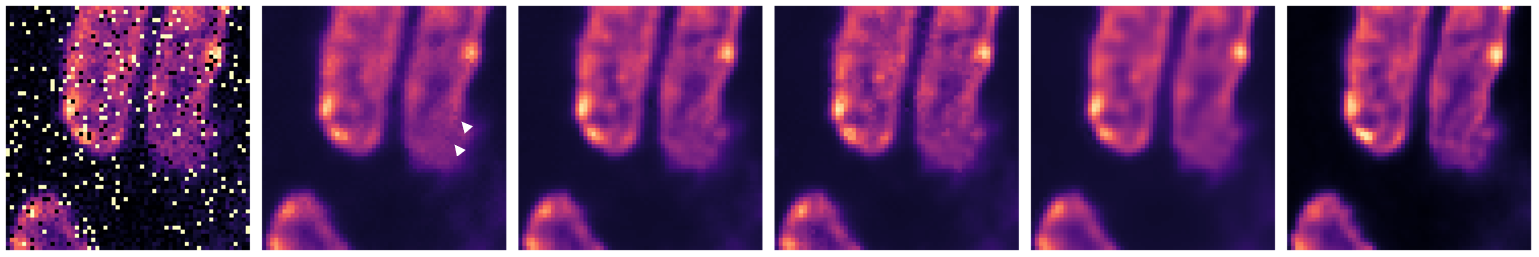}
    \caption{Qualitative results on Mouse SP12 dataset. This data is dominated by 12\% S\&P noise, as can be seen in the input image ($1$\textsuperscript{st}).  The results of the N2V method without residual connection and sampling without the center point ($2$\textsuperscript{nd}, PSNR $33.10$) show checkerboard artifacts, see white arrow heads. 
    On BSD68, these artifacts remain when using median replacement ($4$\textsuperscript{th}, PSNR $32.43$), are reduced in the results when using mean replacement ($3$\textsuperscript{rd}, PSNR $33.01$), and are eliminated in the N2V2 results with median replacement ($5$\textsuperscript{th}, PSNR $33.34$)}
    \label{fig:qualitative_MouseSP12}
\end{figure}

\begin{figure}[tb]
     \centering
     \begin{tabularx}{0.9\textwidth}{Y Y Y Y Y Y Y}
     Input & \shortstack{N2V \\ w/o res\\ w/ \uwoCP}  & \shortstack{N2V\\ w/o res,\\ w/ mean} & \shortstack{N2V\\ w/o res,\\ w/ median} & \shortstack{N2V2\\ w/ median}  & GT \\ 
     \hline
     \end{tabularx}
    \includegraphics[width=0.9\textwidth]{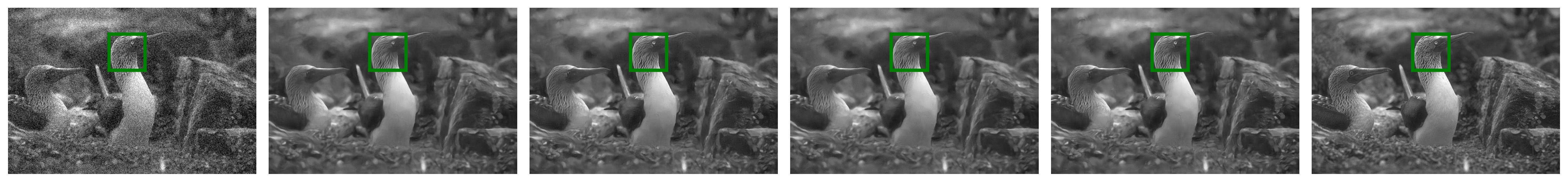}
    \includegraphics[width=0.9\textwidth]{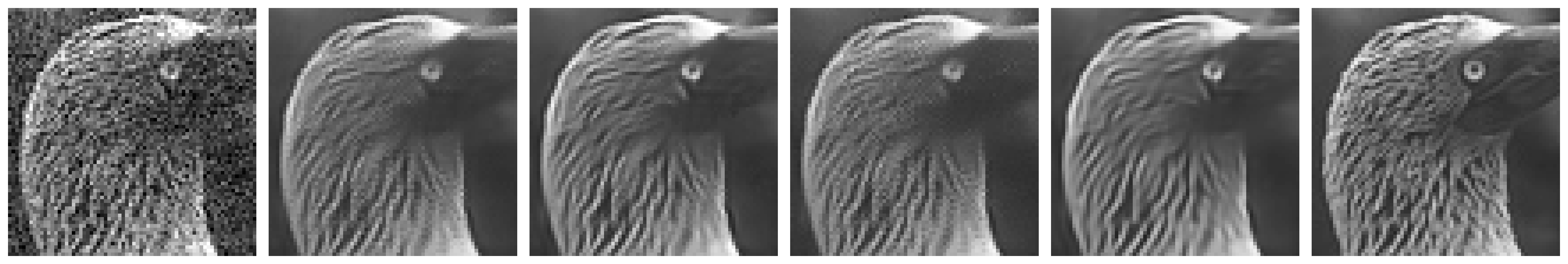}
    \caption{Qualitative results on the BSD68 dataset. After applying the trained N2V model without residual connection and sampling without the center point to the noisy input ($1$\textsuperscript{st}), the result shows undesirable checkerboard artifacts ($2$\textsuperscript{nd}, PSNR $29.01$).
    On BSD68, these artifacts are reduced with mean replacement ($3$\textsuperscript{rd}, PSNR $29.27$), remain when using median replacement ($4$\textsuperscript{th}, PSNR $28.79$),
    and are eliminated when using N2V2 with median replacement ($5$\textsuperscript{th}, PSNR $29.24$)}
    \label{fig:qualitative_BSD}
\end{figure}

\begin{figure}[tb]
     \centering
     \begin{tabularx}{0.9\textwidth}{Y Y Y Y Y Y Y}
     Input & \shortstack{N2V \\ w/o res\\ w/ \uwoCP}  & \shortstack{N2V\\ w/o res,\\ w/ mean} & \shortstack{N2V\\ w/o res,\\ w/ median} & \shortstack{N2V2\\ w/ median}  & GT \\  
     \hline
     \end{tabularx}
    \includegraphics[width=0.9\textwidth]{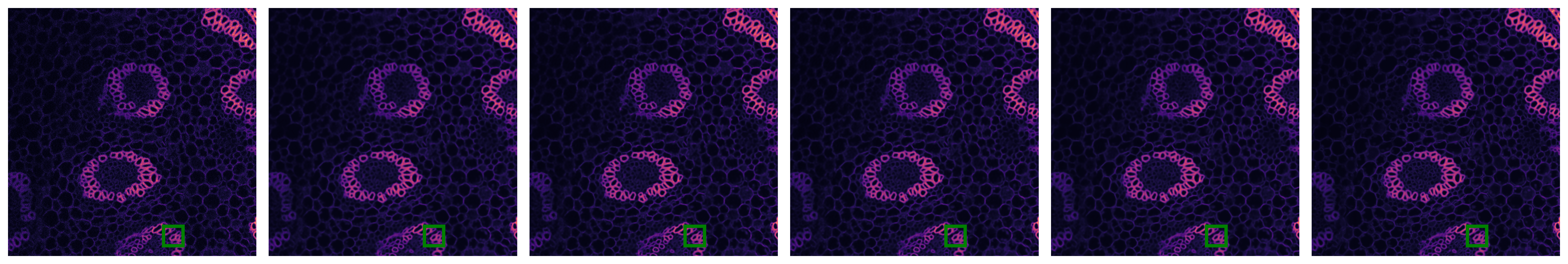}
    \includegraphics[width=0.9\textwidth]{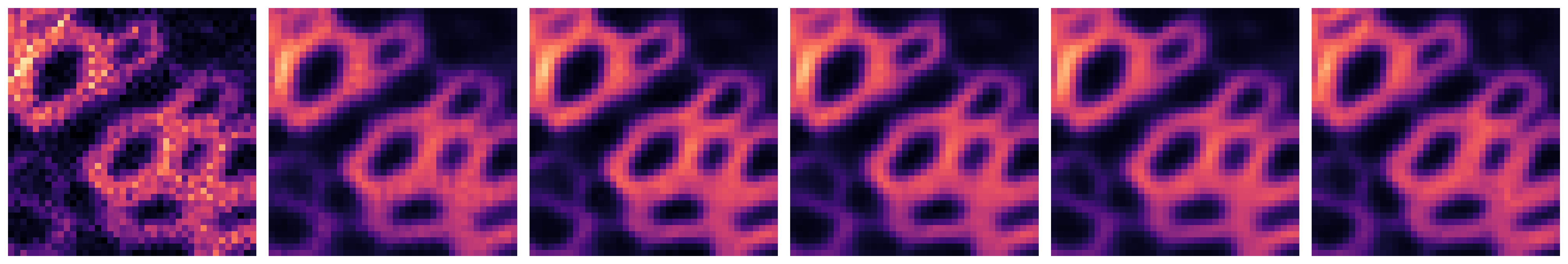}
    \caption{Qualitative results on the convallaria dataset. After applying the trained N2V model without residual connection and sampling without the center point to the noisy input ($1$\textsuperscript{st}), the result shows undesirable checkerboard artifacts ($2$\textsuperscript{nd}, PSNR $35.78$). These are eliminated when using mean ($3$\textsuperscript{rd}, PSNR $35.91$) and median replacement ($4$\textsuperscript{th}, PSNR $36.39$) and as well with the N2V2 with median replacement ($5$\textsuperscript{th}, PSNR $36.37$)}
    \label{fig:qualitative_convallaria}
\end{figure}

We compute PSNR in all conducted experiments, evaluated with respect to the corresponding high-SNR images.
For the BSD68 dataset,
the target range of the PSNR computation is set to $[0, 255]$.
For all other datasets,
the range is obtained by computing the min and max values of each corresponding ground truth image.
We finally report  PSNR values averaged over the entire test data.

\subsection{Results on Mouse SP3, SP6, and SP12 (Salt\&Pepper noise)}
\label{subsec:sp_results}
The results for the S\&P datasets are shown in \Cref{tab:s&presults}. 
First of all,
we see the striking impact of excluding the center pixel from the replacement sampling for S\&P noise: 
while N2V as in \cite{krull2019noise2void} can barely increase the PSNR, 
we see clearly improved results when excluding the center pixel from random sampling for replacement.
In addition, 
a non-residual \UNet further improves the result compared to the residual \UNet that is used by default in the N2V configuration.
In a similar line, also our other architecture adaptations yield increased PSNR values.
 While the proposed replacement strategies mean and median do not result in better quantitative results,
 we are surprised to see that the mean replacement strategy clearly reduces checkerboard artifacts qualitatively as can be seen in \Cref{fig:qualitative_MouseSP12}.
 We finally observe that the best fully self-supervised results in the medium and high noise regime are obtained by combining both architecture and replacement adaptations.

\begin{table}[tb]
    \centering
     \begin{tabular}{l l@{\hskip 12pt} c@{\hskip 12pt} c} 
         \toprule
         & \textbf{Method} & \textbf{Convallaria\_95} & \textbf{Convallaria\_1} \\
         \midrule
          & Input       & $29.40$ & $25.81$  \\
         \hdashline
         \multirow{8}{*}{\shortstack{Fully \\self- \\ supervised}} 
          &  N2V as in \cite{krull2019noise2void}   & $35.89$ & $31.43$  \\
          & N2V w/ \uwoCP as in \cite{pn2v}  & $35.58$ & $31.24$  \\
          &  N2V w/o res, w/ \uwoCP     & $35.76$ & $31.27$ \\
          \cdashline{2-4}
          & N2V w/o res  w/ mean       & $35.90$ & $31.34$ \\
          & N2V w/o res  w/ median     & $\mathbf{36.39}$ & $\mathbf{31.77}$ \\
          & N2V2 w/ \uwCP & $36.26$ & $31.45$ \\
          & N2V2 w/ \uwoCP & $36.31$ & $31.51$ \\
          & N2V2 w/ mean & $36.27$ & $31.48$ \\
          & N2V2 w/ median & $36.36$ & $31.28$ \\
         \hdashline
         \multirow{2}{*}{\shortstack{Self-\\ supervised}} & PN2V \cite{pn2v} & $36.47$ & N/A \\
          & DivNoising \cite{divnoising} & $\underline{36.90}$ & N/A \\
         \hdashline
         Supervised & CARE \cite{weigert2018content} & $36.71$ & N/A \\
         \bottomrule
     \end{tabular}
     \caption{Quantitative results on real Convallaria data. Results are given in dB of averaged PSNR on test data. Overall best is $\text{\underline{underlined}}$. Best fully self-supervised is in \textbf{bold}}
     \label{tab:convallariaresults}
\end{table}

\begin{table}[tb]
    \centering
     \begin{tabular}{l l@{\hskip 12pt} c@{\hskip 12pt} c} 
         \toprule
         & \textbf{Method} & \textbf{Convallaria\_1 train} & \textbf{Convallaria\_1 test} \\
         \midrule
          & Input                         & $25.21$ & $25.81$  \\
         \hdashline
         \multirow{5}{*}{\shortstack{Fully \\self- \\ supervised}} 
         &    N2V w/o res, w/ \uwoCP      & $30.52$ & $31.27$ \\
          \cdashline{2-4}
          & N2V w/o res,  w/ mean         & $31.37$ & $31.34$ \\
          & N2V w/o res,  w/ median       & $31.35$ & $31.77$ \\
          & N2V2 w/ mean                  & $31.10$ & $31.48$ \\
          & N2V2 w/ median                & $31.06$ & $31.28$ \\
         \bottomrule
     \end{tabular}
     \caption{Quantitative results on the Convallaria\_1 train and test sets. Results are given in dB of averaged PSNR}
     \label{tab:convallariaresults_train}
\end{table}

\subsection{Evaluation Flywing G70, Mouse G20, BSD68}
\label{subsec:results_gaussian_noise}
We report results for the datasets with simulated Gaussian noise in \Cref{tab:gaussiannoiseresults}. 
In contrast to the results for simulated salt and pepper noise,
we interestingly see that results do not improve simply by excluding the center pixel from the window for sampling replacement.
Also, not using a residual U-Net only yields slight improvements for the microscopy datasets and none for the natural image dataset BSD68, where PSNR even drops.
However,
the alternative replacement strategies mean and median lead to improved PSNR values, as well as the architecture adaptations bp sk.
Combining both adaptations leads to the best self-supervised results for the Mouse G20 and BSD68 datasets.

This is in line with qualitative results shown in \Cref{fig:qualitative_BSD} for the BSD68 dataset, where we clearly see checkerboard artifacts in the N2V standard setting, but significantly cleaner predictions with the proposed adaptations.
Additional qualitative results are given in the supplementary material in section S.1.

\subsection{Evaluation of Real Noisy Data: Convallaria\_95 and Convallaria\_1}
\label{subsec:results_convallaria}
As displayed in \Cref{tab:convallariaresults}, both the median replacement strategy as well as the N2V2 architecture adaptations improve the results for both Convallaria datasets. 
This can also be seen in the qualitative example in \Cref{fig:qualitative_convallaria}.
N2V2 with median replacement strategy yields the best fully self-supervised results for both cases. 
Interestingly, according to PSNR values, the mean replacement method does not improve when compared to the baseline N2V performance.

\begin{figure}[b]
    \centering
    \includegraphics[width=0.7\textwidth]{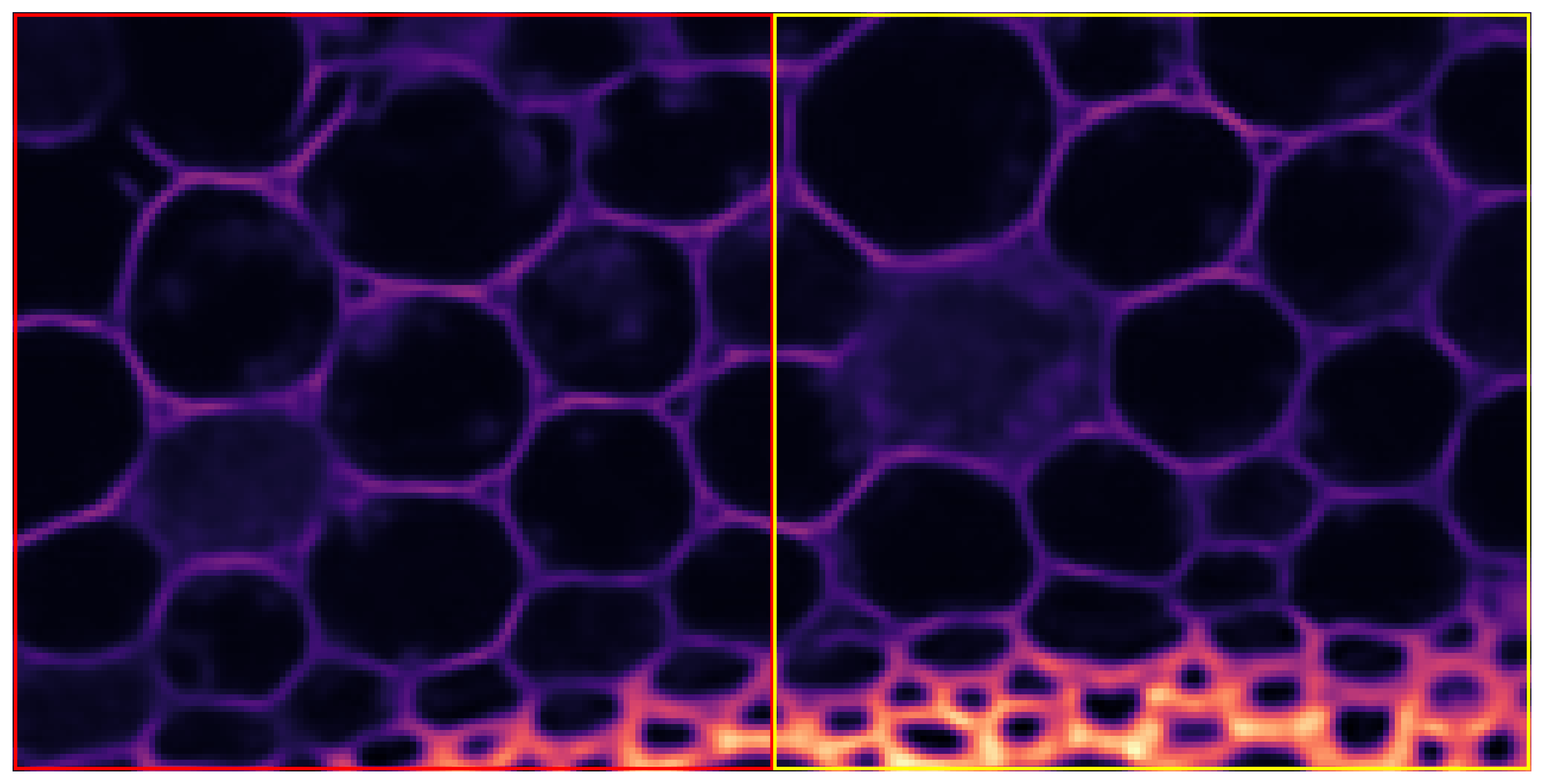}
    \caption{Does prediction on training data impact N2V quality? Predictions on data which was used for training N2V (left) and hold-out data (right)}
    \label{fig:predict_on_train_or_test}
\end{figure}

Comparing the two columns in Table \ref{tab:convallariaresults}, a considerable difference in PSNR is apparent, with the denoising results when using the reduced Convallaria\_1 dataset being poorer. 
This leads to two possible interpretations, namely
$(i)$~having 95 noisy images of the same field of view allows for better results of the self-supervised denoising methods or
$(ii)$~results are poorer on the hold-out tiles of the Convallaria\_1 test set because they represent parts of the field of view that were not seen during training. 
However, judging by \Cref{tab:convallariaresults_train}, which displays a comparison of the results on the train vs. the test tiles, this seems not to be the case. 
A similar conclusion is suggested by \Cref{fig:predict_on_train_or_test}, showing a qualitative comparison between denoised train and test tiles.
Please also refer to the supplementary material section S.2 for additional qualitative results obtained for the whole slide.

\section{Discussion \& Conclusions}
\label{sec:discussion}
In this work, we introduced N2V2, an improved setup for the self-supervised denoising method N2V by Krull \etal~\cite{krull2019noise2void}.
N2V2 is build around two complementary contributions:
$(i)$~a new network architecture, and 
$(ii)$~modified pixel value replacement strategies for blind-spot pixels.

We showed that N2V2 reduces previously observed checkerboard artifacts,
which have been responsible for reduced quality of predictions from N2V. 
While we observed in qualitative examples that the mean replacement strategy is overall more successful than the median replacement strategy, we did not find this trend consistently in all quantitative results.
Nonetheless, we have shown that only changing the architecture or only switching to one of our sampling strategies does already lead to improved results.
Still, the combination of both yields best overall denoising results (measured by means of PSNR to ground clean truth images).

An interesting observation regards the failed denoising of N2V with uwCP sampling, \ie the original N2V base line, in the S\&P noise setting. The network learns only to remove the Gaussian noise and the pepper noise, but recreates the salt noise pixels in the prediction (see Supplementary Section S.3). We attributed this to the strong contrast of the salt pixels with respect to the rest of the image, and the probability of $\frac{1}{w'\cdot h'}$ that the pixel remains unchanged allowing the network to learn the identity for such pixels.
Another commonly used default in N2V denoising might be hurting the performance:  When using a residual \UNet, pixels altered by a huge amount of noise appear at times to be strongly biased by the residual input and denoising is therefore negatively effected, as can be seen from Table \ref{tab:s&presults}. 
Without residual connections, on the other hand, this bias is removed and  performance therefore improved.

Additionally, we have introduced a modified Convallaria data set (\textit{Convallaria\_1}), now featuring 
$(i)$~a clean split between train, validation, and test sets, and offering
$(ii)$~a more realistic scenario to test self-supervised denoising methods. 
The newly proposed dataset includes only one noisy input image instead of the previously used $99$ noisy acquisitions of the same field of view of the same sample.
We strongly urge the community to evaluate future methods on this improved Convallaria setup.

As a final point of discussion, we note that since we decided to train all N2V and N2V2 setups much longer than in previous publications (\eg \cite{pn2v}), even the baselines we have simply re-run now outperform the corresponding results as reported in the respective original publications.
This indicates that original training times were chosen too short and urges all future users of self-supervised denoising methods to ensure that their training runs have indeed converged before stopping them\footnote{Note that this is harder to judge for self-supervised compared to supervised methods since loss plots report numbers that are computed between predicted values and \textit{noisy} blind-spot pixel values.}. 

With N2V2, we presented an improved version of N2V, a self-supervised denoising method leading to denoising results of improved quality on virtually all biomedical microscopy data. 
At the same time, N2V2 is equally elegant, does not require more or additional training data, and is equally computationally efficient as N2V.
Hence, we hope that N2V2 will mark an important update of N2V and will continue the success which N2V has celebrated in the past three years.

\subsection*{Acknowledgements}
The authors would like to thank Laurent Gelman of the Facility for Advanced Imaging and Microscopy (FAIM) at the FMI for biomedical research in Basel to provide resources to perform the reported experiments.

\FloatBarrier

\bibliographystyle{splncs04}
\bibliography{0196}


\newpage

\setcounter{equation}{0}
\setcounter{figure}{0}
\setcounter{table}{0}
\setcounter{page}{1}
\renewcommand{\thepage}{S.\arabic{page}} 
\renewcommand{\thefigure}{S.\arabic{figure}}
\renewcommand*{\thesection}{S.\arabic{section}}
\renewcommand{\titlerunning}{Supp.: N2V2 - Fixing Noise2Void Checkerboard Artifacts}
\setcounter{section}{0}

\begin{center}
  \textbf{\large 
  N2V2 - Fixing Noise2Void Checkerboard Artifacts with Modified Sampling Strategies and a Tweaked Network Architecture \\ -- Supplementary Material -- }\\[0.1cm]
Eva H{\"o}ck$^{1,\ast}$ \orcidZEISS{0000-0003-2133-1282}, %
Tim-Oliver Buchholz$^{2,\ast}$\orcidFMI{0000-0001-6953-8915},  %
Anselm Brachmann$^{1,\ast}$, \\
Florian~Jug$^{3,\circledast}$\orcidHT{0000-0002-8499-5812}, 
Alexander~Freytag$^{1,\circledast}$\orcidZEISS{0000-0002-9041-1334}\\[.1cm]
\blfootnote{$^\ast$, $^\circledast$ Equal contribution}
\textsuperscript{1}Carl Zeiss AG, Germany \\
\textsuperscript{2}Facility for Advanced Imaging and Microscopy, Friedrich Miescher Biomedical Research, Basel, Switzerland \\
\textsuperscript{3} Jug Group, Fondazione Human Technopole, Milano, Italy
\\
\texttt{eva.hoeck@zeiss.com, 
tim-oliver.buchholz@fmi.ch, 
anselm.brachmann@zeiss.com,
florian.jug@fht.org,
alexander.freytag@zeiss.com}
\end{center}

\begin{abstract}
In this supplementary document,
we provide additional qualitative results to further strengthen our findings as reported in the main paper.
\Cref{supplmat:BSD_qualitative} contains additional qualitative results on the BSD68 dataset.
\Cref{supplmat:Convallaria_qualitative} shows whole-slide results on the Convallaria\_1 dataset.
Finally, \Cref{sec:N2V_and_SP_noise} shows a finding when using plain N2V in the presence of strong salt-and-pepper noise.
\end{abstract}

\section{Additional qualitative results for BSD68}
\label{supplmat:BSD_qualitative}
\begin{figure}[!htb]
     \centering
     \begin{tabularx}{0.9\textwidth}{Y Y Y Y Y Y}
     Input & \shortstack{N2V \\ w/o res\\ w/ \uwoCP}  & \shortstack{N2V\\ w/o res,\\ w/ mean} & \shortstack{N2V\\ w/o res,\\ w/ median} & \shortstack{N2V2\\ w/ median}  & GT \\
     \hline
     \end{tabularx}
    \includegraphics[width=0.9\textwidth]{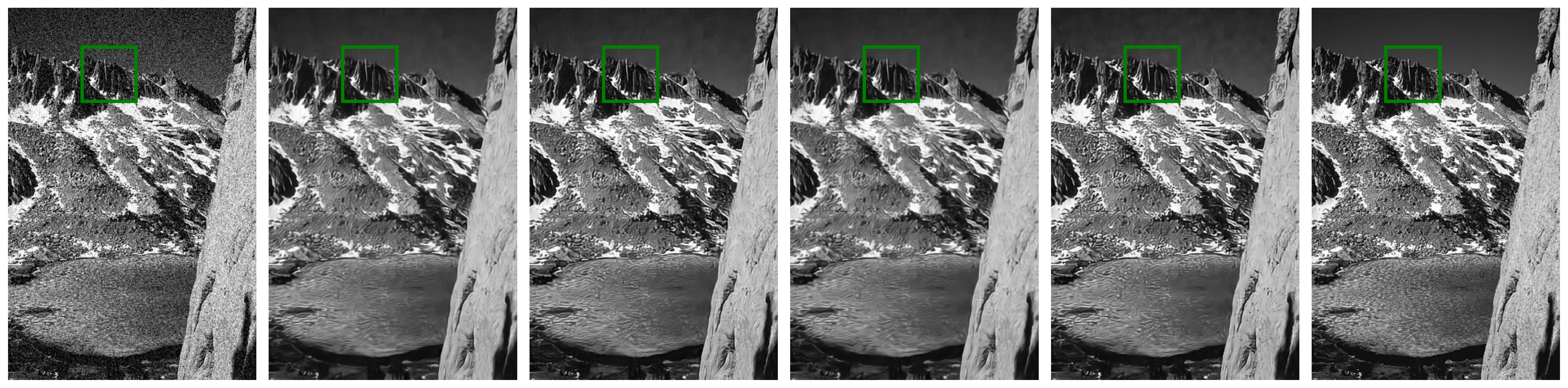}
    \includegraphics[width=0.9\textwidth]{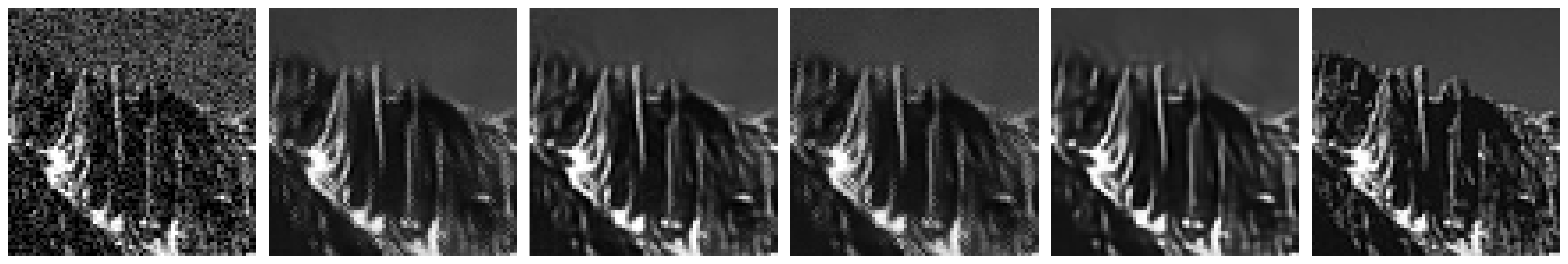}
    \includegraphics[width=0.9\textwidth]{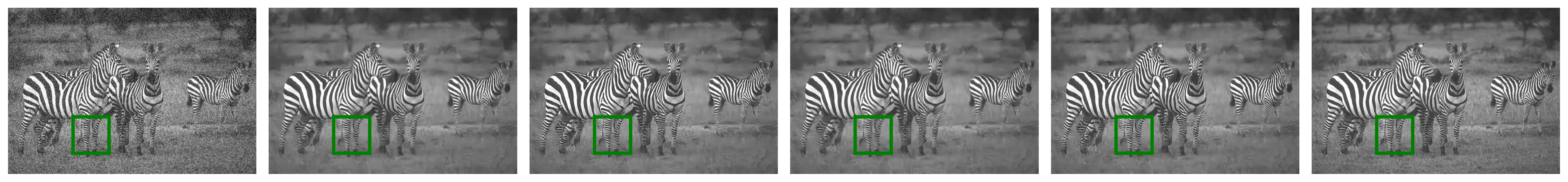}
    \includegraphics[width=0.9\textwidth]{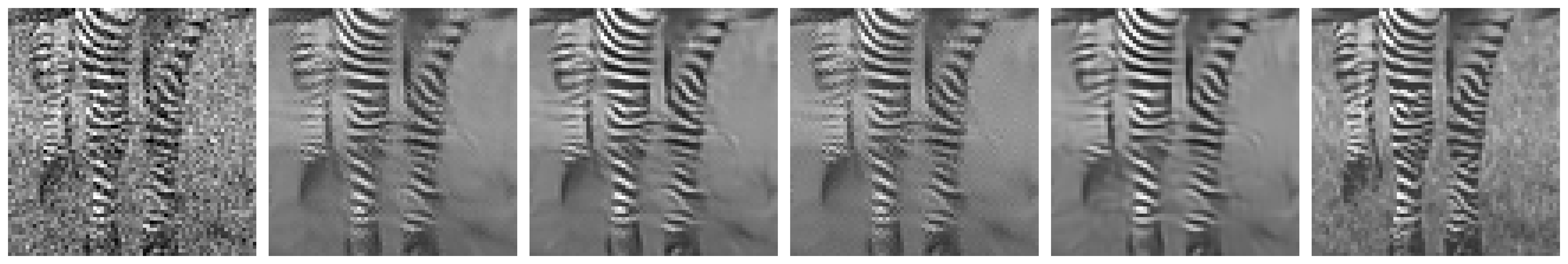}
    \includegraphics[width=0.9\textwidth]{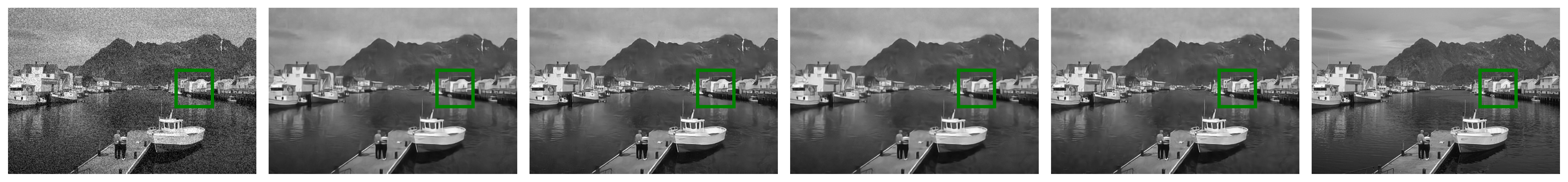}
    \includegraphics[width=0.9\textwidth]{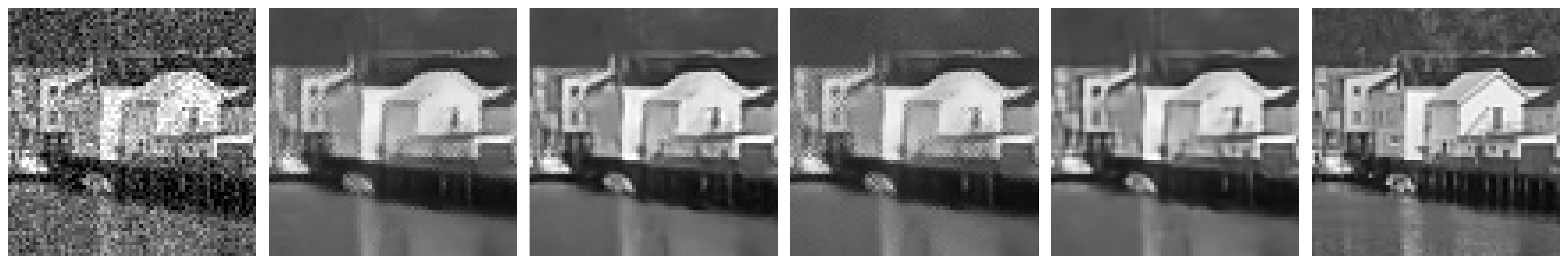}
    \includegraphics[width=0.9\textwidth]{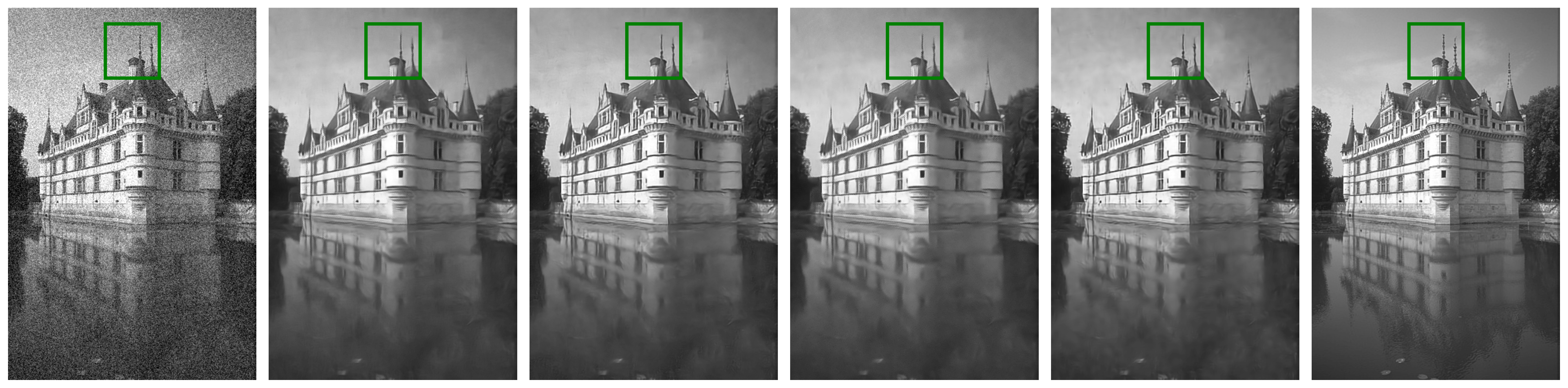}
    \includegraphics[width=0.9\textwidth]{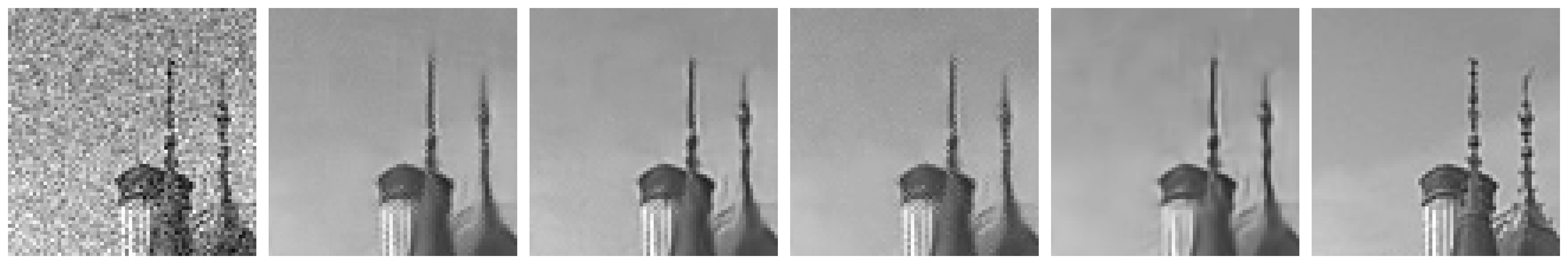}
    \caption{Additional qualitative results on the BSD68 dataset. After applying the trained N2V model without residual connection and sampling without the center point to the noisy input ($1$\textsuperscript{st}), the result shows undesirable checkerboard artifacts ($2$\textsuperscript{nd}).
     On BSD68, these artifacts are reduced with mean replacement ($3$\textsuperscript{rd}), remain when using median replacement ($4$\textsuperscript{th}),
    and are eliminated when using N2V2 with median replacement ($5$\textsuperscript{th})}
    \label{fig:supplmat_qualitative_BSD}
\end{figure}

\begin{figure}[!htb]
     \centering
     \begin{tabularx}{0.9\textwidth}{Y Y Y Y Y Y Y}
     Input & \shortstack{N2V \\ w/o res\\ w/ \uwoCP}  & \shortstack{N2V\\ w/o res,\\ w/ mean} & \shortstack{N2V\\ w/o res,\\ w/ median} & \shortstack{N2V2\\ w/ median}  & GT \\ 
     \hline
     \end{tabularx}
    \includegraphics[width=0.9\textwidth]{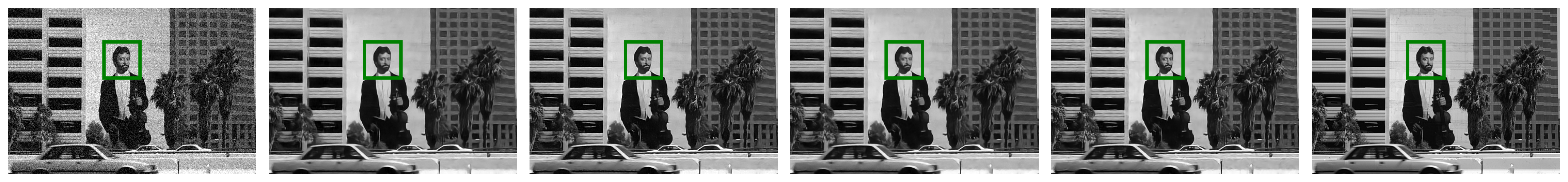}
    \includegraphics[width=0.9\textwidth]{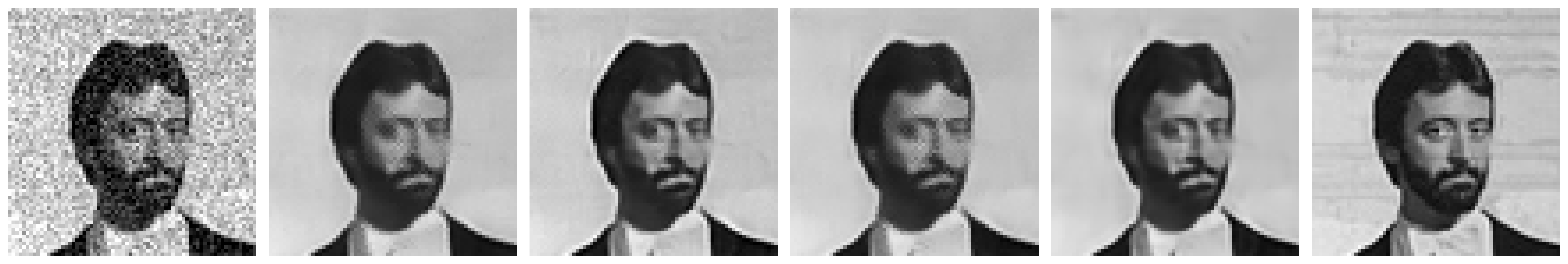}
    \includegraphics[width=0.9\textwidth]{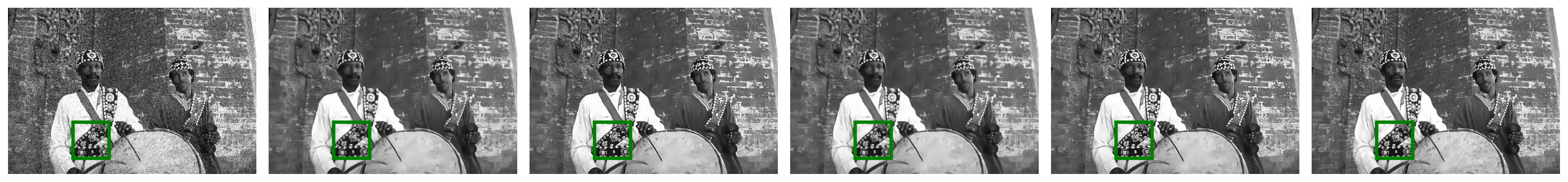}
    \includegraphics[width=0.9\textwidth]{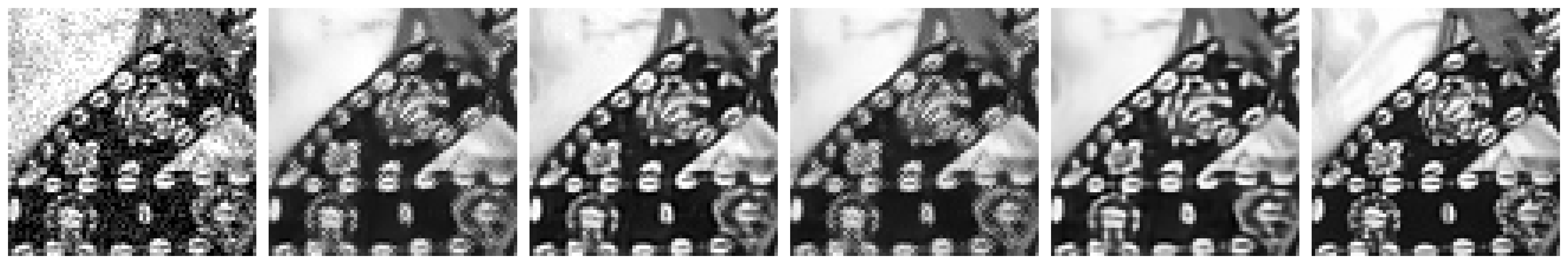}
    \includegraphics[width=0.9\textwidth]{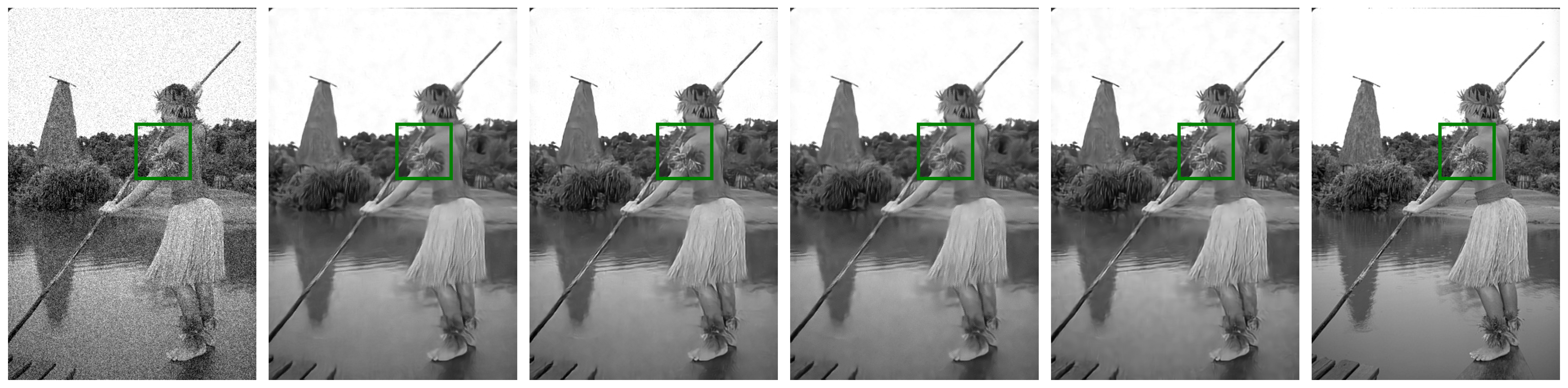}
    \includegraphics[width=0.9\textwidth]{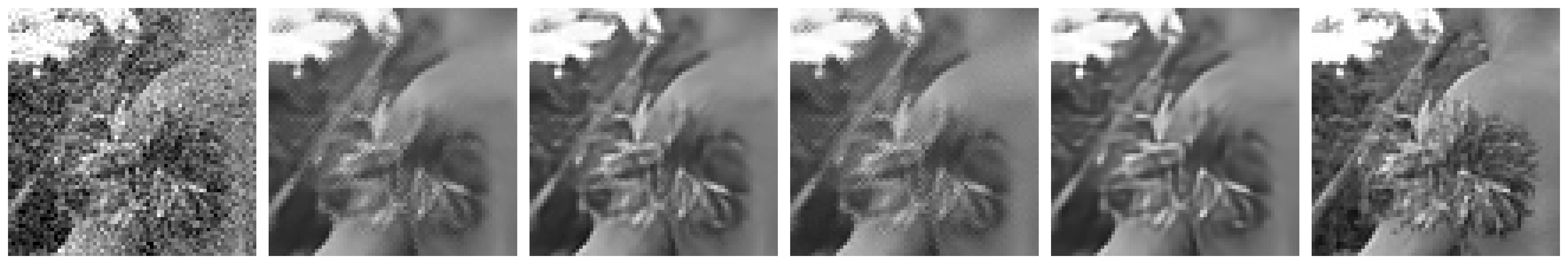}
    \includegraphics[width=0.9\textwidth]{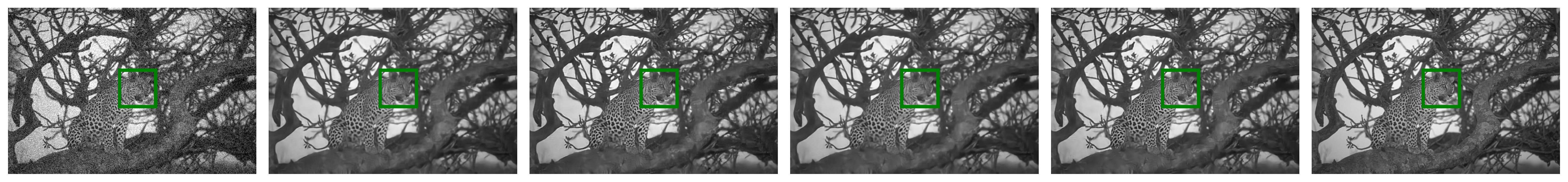}
    \includegraphics[width=0.9\textwidth]{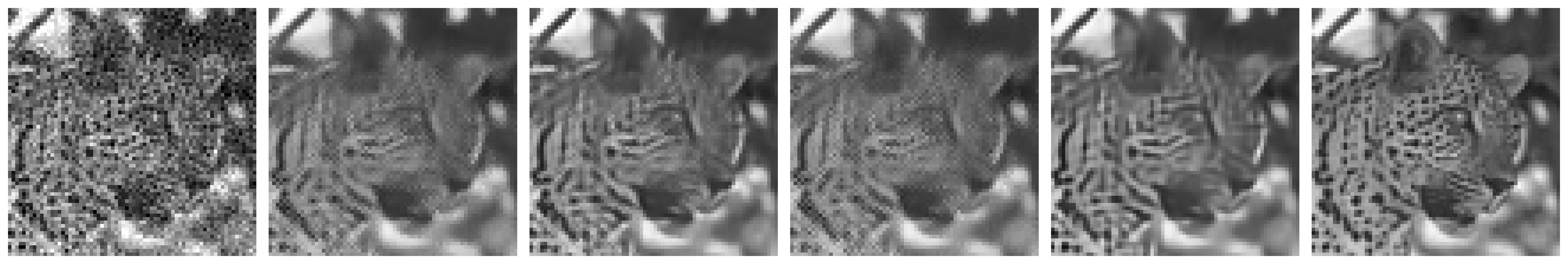}
    \caption{Even more qualitative results on the BSD68 dataset. After applying the trained N2V model without residual connection and sampling without the center point to the noisy input ($1$\textsuperscript{st}), the result shows undesirable checkerboard artifacts ($2$\textsuperscript{nd}).
    On BSD68, these artifacts are reduced with mean replacement ($3$\textsuperscript{rd}), remain when using median replacement ($4$\textsuperscript{th}),
    and are eliminated when using N2V2 with median replacement ($5$\textsuperscript{th})}
    \label{fig:supplmat_qualitative_BSD2}
\end{figure}

In the main paper,
we report quantitative and qualitative results in Section 4.4 for the BSD68 natural images dataset.
In \Cref{fig:supplmat_qualitative_BSD} and \Cref{fig:supplmat_qualitative_BSD2}, 
we add more qualitative results to further underline the benefits of N2V2.

\section{Whole slide results on the Convallaria\_1 dataset}
\label{supplmat:Convallaria_qualitative}
\begin{figure}[!htb]
    \centering
    \includegraphics[width=0.9\textwidth]{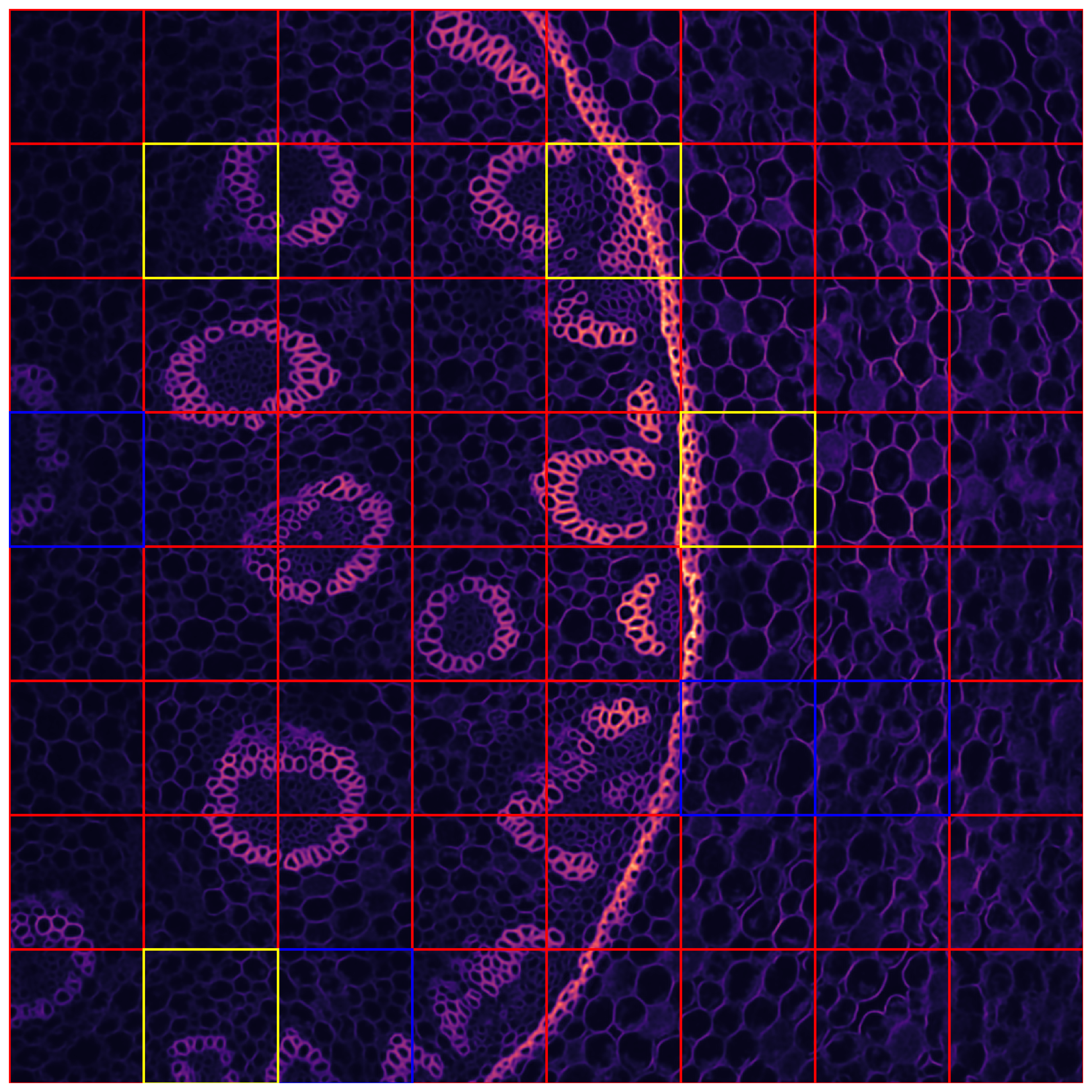}
    \caption{What is used for train and test? \textcolor{red}{Red}: used for training. \textcolor{blue}{Blue}: used for validation (\ie, not for network parameter optimization during training of N2V, but we still create a val-set due to historical reasons). \textcolor{orange}{Yellow}: only used for testing}
    \label{suppfig:full_slide_prediction}
\end{figure}

In the main paper, 
we report results on the Convallaria\_1 dataset in Section 4.5.
We discuss that although a clear separation of training data and test data is  a sound experimental setup even for self-supervised training scenarios, no clear differences for denoising results have been observed.
In \Cref{suppfig:full_slide_prediction},
we show additional qualitative results obtained with N2V2 w/ mean by visualizing denoising results on the entire Convallaria slide in Figure~\ref{suppfig:full_slide_prediction}.
The origin of each patch, \ie if being used in the training set, validation set, or test set, is further indicated by the colored frame.

\section{Noise2Void with uwCP replacement on Salt \& Pepper noise}
\label{sec:N2V_and_SP_noise}
\begin{figure}[!h]
    \centering
    \includegraphics[width=0.3\textwidth]{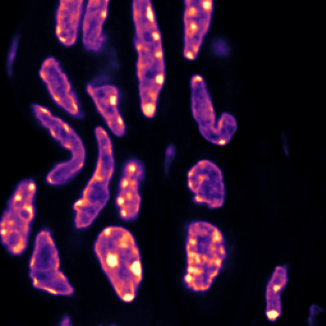}
    \includegraphics[width=0.3\textwidth]{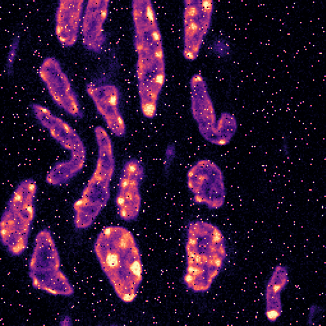}
    \includegraphics[width=0.3\textwidth]{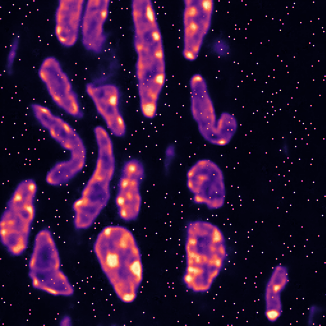}
    \caption{\textit{Left}: the clean ground truth image. \textit{Middle}: the input degraded with Poisson, Gaussian and S\&P noise (SP3). \textit{Right}: the denoised output of standard N2V trained with uwCP pixel replacement. Note the presence of Salt-pixels even in the denoised output}
    \label{suppfig:n2v_uwCP_sp}
\end{figure}

We finally report an interesting observation with regard to the failed denoising of N2V with uwCP sampling, \ie the original N2V baseline, in the S\&P noise setting.
As can be seen in \Cref{suppfig:n2v_uwCP_sp}, the network learns only to remove the Gaussian noise and the pepper noise, but recreates the salt noise pixels in the prediction. 
We attribute this to the strong contrast of the salt pixels with respect to the rest of the image, and the probability of $\frac{1}{w'\cdot h'}$ that the pixel remains unchanged allowing the network to learn the identity for such pixels.

\end{document}